\documentclass[letterpaper, 10 pt, conference]{ieeeconf}  

\IEEEoverridecommandlockouts                              

\usepackage{graphics} 
\usepackage{epsfig} 

\usepackage{amsmath,amssymb,setspace,fancyhdr,dsfont,graphicx,hyperref,bbm}

\usepackage{subcaption}

\usepackage[space]{cite}

\title{\LARGE \bf
Electroadhesive Auxetics as Programmable Layer Jamming Skins for Formable Crust Shape Displays
}

\author{Ahad M. Rauf$^{1}$, Jack S. Bernardo$^{1}$, and Sean Follmer$^{1}$
\thanks{$^{1}$Authors are with the Department of Mechanical Engineering at Stanford University, Stanford, CA 94305, USA.
        {\tt\small ahadrauf@stanford.edu}}
}

\begin{document}

\maketitle
\thispagestyle{empty}
\pagestyle{empty}

\begin{abstract}

Shape displays are a class of haptic devices that enable whole-hand haptic exploration of 3D surfaces. However, their scalability is limited by the mechanical complexity and high cost of traditional actuator arrays. In this paper, we propose using electroadhesive auxetic skins as a strain-limiting layer to create programmable shape change in a continuous (``formable crust'') shape display. Auxetic skins are manufactured as flexible printed circuit boards with dielectric-laminated electrodes on each auxetic unit cell (AUC), using monolithic fabrication to lower cost and assembly time. By layering multiple sheets and applying a voltage between electrodes on subsequent layers, electroadhesion locks individual AUCs, achieving a maximum in-plane stiffness variation of 7.6x with a power consumption of 50 \textmu W/AUC. We first characterize an individual AUC and compare results to a kinematic model. We then validate the ability of a 5x5 AUC array to actively modify its own axial and transverse stiffness. Finally, we demonstrate this array in a continuous shape display as a strain-limiting skin to programmatically modulate the shape output of an inflatable LDPE pouch. Integrating electroadhesion with auxetics enables new capabilities for scalable, low-profile, and low-power control of flexible robotic systems.
\end{abstract}

\section{Introduction and Related Work} \label{sec:introduction}


Shape displays allow users to freely explore and manipulate surfaces through haptic interactions, a concept known as ``digital clay'' \cite{Rossignac_Allen_Book_Glezer_Ebert-Uphoff_Shaw_Rosen_Askins_Bai_Bosscher_2003, Bosscher_Ebert-Uphoff_2003}. They traditionally work using an array of linear actuators that move up and down to render a discretized 2.5D depth map \cite{Follmer_Leithinger_Olwal_Hogge_Ishii_2013, Siu_Gonzalez_Yuan_Ginsberg_Follmer_2018}. Shape displays are powerful tools for design and architecture, enabling tactile manipulation of CAD models, physical user interfaces that can adapt to individual users or tasks, and accessible 3D prototyping tools for blind or visually impaired makers \cite{Alexander_Roudaut_Steimle_Hornbaek_Bruns_Alonso_Follmer_Merritt_2018, Siu_Kim_Miele_Follmer_2019}. However, discretized actuator arrays scale poorly to smooth surfaces, where their mechanical complexity and cost inhibit immersive render resolutions and display sizes \cite{Steed_Ofek_Sinclair_Gonzalez-Franco_2021, Abtahi_Follmer_2018}.



Formable crust shape displays differ from conventional motor arrays by embedding  the actuation mechanism into the surface itself,  reducing  assembly bulk and facilitating curved surface rendering \cite{Rossignac_Allen_Book_Glezer_Ebert-Uphoff_Shaw_Rosen_Askins_Bai_Bosscher_2003, Bosscher_Ebert-Uphoff_2003}. Using elastomeric materials as flexible substrates, several actuation methods have been tested for these programmable surfaces, including pneumatics \cite{Stanley_Okamura_2015, Shah_Yang_Yuen_Huang_Kramer_Bottiglio_2021}, hydraulics \cite{Rossignac_Allen_Book_Glezer_Ebert_Uphoff_Shaw_Rosen_Askins_Bai_Bosscher_2003}, and shape memory alloys \cite{Coelho_Ishii_Maes_2008}. However, these typically require complicated and expensive fabrication, long actuation times, or  high power consumption, impeding large-scale continuous shape displays.

\begin{figure}
    \centering
    \includegraphics[width=\columnwidth,trim={2pt 0 0 0},clip]{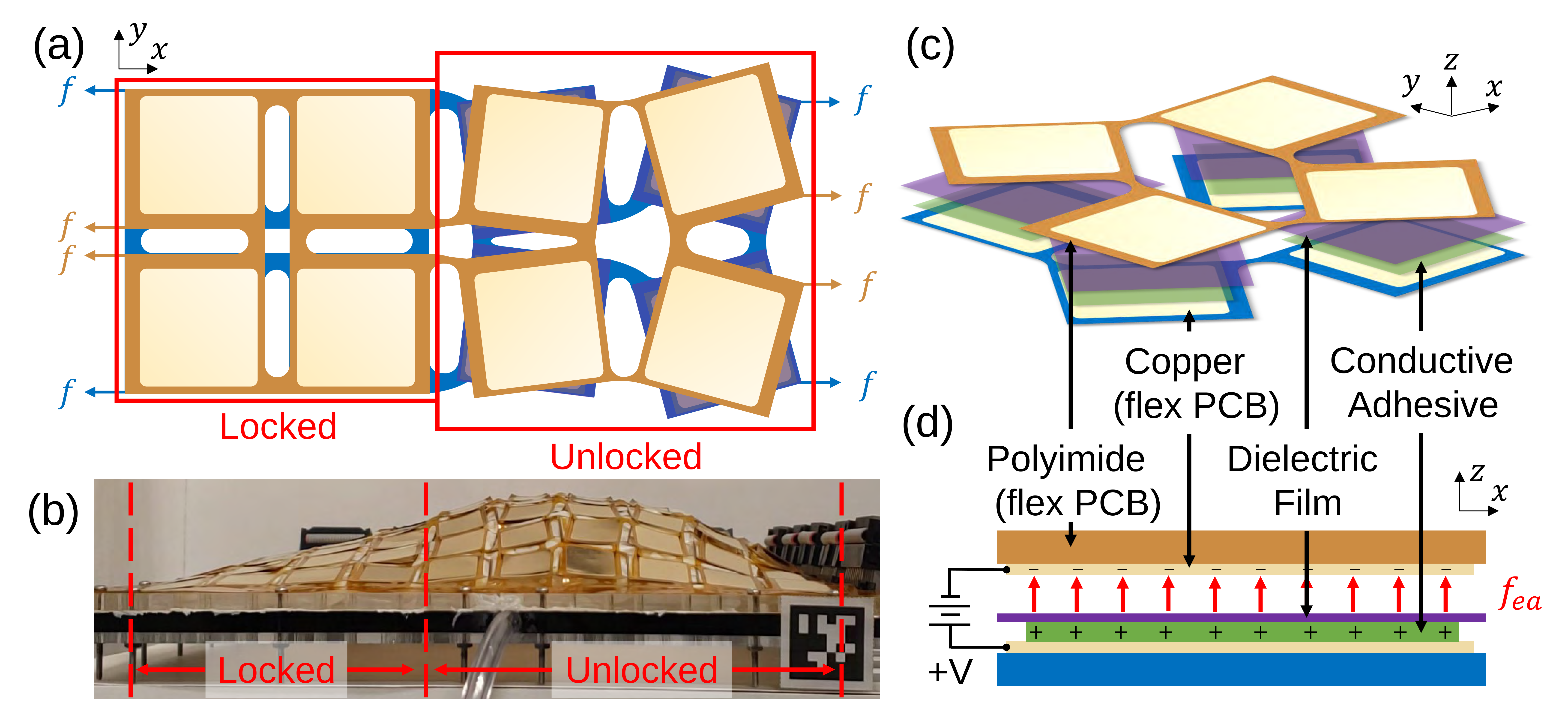}
    \caption{Conceptual diagram for a 2-layer auxetic sheet stack. The bottom sheet's auxetic unit cells (AUCs, blue) are rotated by 90$^\circ$ relative to the top sheet's AUCs (orange), causing their squares to rotate in opposite directions when stretched. (a) Auxetic layer stack stretched axially by force $4f$. The left AUC is locked in place via electroadhesion, while the right AUC is free to expand. (b) Locking only part of a formable crust shape display's surface results in global shape change. (c) Exploded view of the electroadhesive layer stack for a single AUC. (d) Electroadhesive normal force $f_{ea}$ is generated between the metallized electrodes and amplified by the dielectric film when a voltage differential is applied.}
    \label{fig:conceptual_diagram}
\end{figure}

Auxetics present a promising alternative to elastomeric materials when integrating programmable-stiffness capabilities into flexible surfaces. Auxetics have a negative Poisson's ratio ($\nu < 0$), allowing them to expand and conform to high-curvature surfaces without collapsing \cite{Sedal_Memar_Liu_Menguc_Corson_2020, Konakovic-Lukovic_Panetta_Crane_Pauly_2018}.  They can be  made from traditional materials and plastics using monolithic manufacturing techniques like flexible printed circuit board (PCB) fabrication and laser cutting. This lowers assembly complexity and cost while integrating electronics for programmable actuation \cite{Jang_Won_Kim_Kim_Oh_Lee_Kim_2022}. Prior works  using 3D printed \cite{Xin_Liu_Liu_Leng_2020, An_Tao_Gu_Cheng_Chen_Zhang_Zhao_Do_Takahashi_Wu_2018} or laser cut \cite{Sedal_Memar_Liu_Menguc_Corson_2020, Konakovic-Lukovic_Panetta_Crane_Pauly_2018, Rafsanjani_Zhang_Liu_Rubinstein_Bertoldi_2018} meta-materials  have applied these properties to create passive auxetic skins with pre-programmed stiffness profiles for soft robots and shape displays, allowing a single device to produce a variety of output shapes by manually changing the skin wrapped around it at the time. However, little work has embedded programmable stiffness capabilities into  
planar  auxetic materials to modulate the shape output at runtime.  We propose that by fabricating auxetic sheets as flexible PCBs, we can integrate planar electroadhesion actuation with auxetics for scalable, low-cost, and low-power on-demand stiffness control.

Electroadhesion is the electrostatic attractive force between two metallized surfaces separated by a dielectric material and held at a potential difference. It has gained increasing interest for its high force output given its low mass and power consumption \cite{Guo_Leng_Rossiter_2020}, with applications including low-profile clutches \cite{Zhang_Gonzalez_Guo_Follmer_2019, Park_Drew_Follmer_Rivas-Davila_2020, Ramachandran_Shintake_Floreano_2019}, exoskeleton actuation \cite{Diller_Collins_Majidi_2018, Hinchet_Vechev_Shea_Hilliges_2018}, and conformal grippers \cite{Aukes_Heyneman_Ulmen_Stuart_Cutkosky_Kim_Garcia_Edsinger_2014, Han_Hajj-Ahmad_Cutkosky_2020, Seitz_Goldberg_Doshi_Ozcan_Christensen_Hawkes_Cutkosky_Wood_2014}.  Its integration into volumetric auxetic structures \cite{Heath_Neville_Scarpa_Bond_Potter_2016} shows promise for tunable stiffness control, motivating our extension to planar auxetic surfaces.

Fig. \ref{fig:conceptual_diagram} shows the layer stack we used to integrate electroadhesion with auxetic sheets, for use as a strain limiting layer in a formable crust shape display. We fabricated auxetic sheets as flexible PCBs with metallized electrodes on each AUC, and one sheet (colored blue in Fig. \ref{fig:conceptual_diagram}) is assembled with conductive adhesive and a dielectric film to amplify the electrostatic attractive force. When another auxetic sheet (colored orange) with exposed metallized electrodes is rotated by 90$^\circ$ and laid over the assembled sheet, their squares rotate in opposite directions when stretched. Applying a voltage between the two layers generates an electroadhesive normal force and thus a frictional shear braking torque, locking the AUC and preventing it from expanding. This layer stack can be extended to three or more auxetic sheets by metallizing both sides of the flexible PCB and alternating high voltage and ground layers. Fig. \ref{fig:conceptual_diagram}(b) demonstrates how locally locking selected AUCs results in global shape change when overlaid as a strain limiting layer on top of an inflatable  low-density polyethylene (LDPE)  pouch. Low-stiffness regions inflate more given an isotropic pressure input, achieving higher curvature than locked surface regions.

In this work, we motivate the integration of electroadhesion as a programmable stiffness actuation mechanism for auxetic sheets. We characterize these stiffening properties across a variety of materials and match the experimental data to a kinematic model with good conformity. We explore the ability of locking multiple AUCs in an auxetic array to control the sheets' axial and transverse stiffness, achieving global shape change when stretched. Finally, we demonstrate the ability of this actuator to modulate the shape output of a continuous, formable crust shape display.  A video of the device in action can be found at \url{https://youtu.be/c7cHUNBqbKU}.

\section{Design of an Individual Electroadhesive Auxetic Unit Cell Layer Stack} \label{sec:characterization_2x2}

\begin{figure}
    \centering
    \includegraphics[width=\columnwidth,trim={10pt 10pt 0 0},clip]{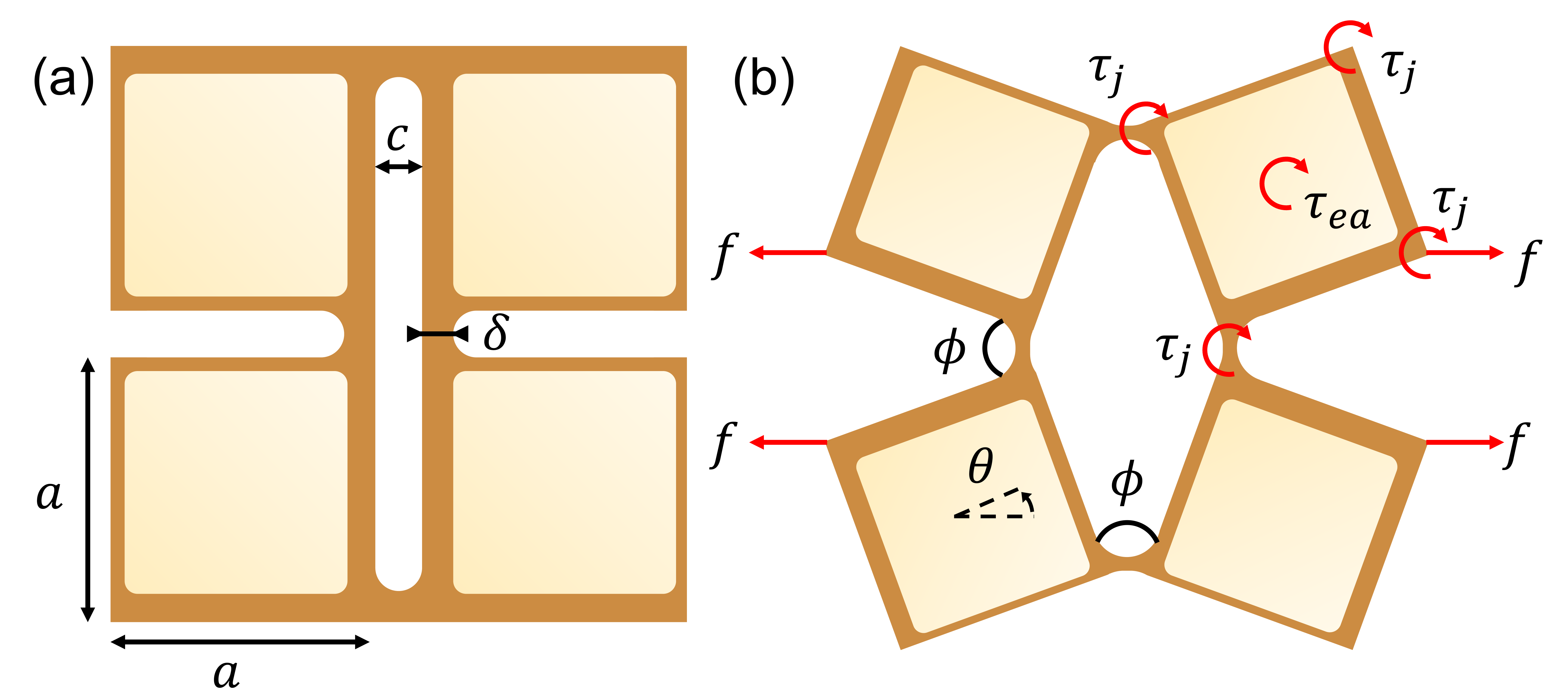}
    \caption{Dimensional drawing of square AUC when (a) unstretched and (b) stretched. The metallized pad is offset by 0.444mm to account for space taken up by electrical wiring.}
    \label{fig:kinematics_dimension_drawing}
\end{figure}

\subsection{Planar Kinematic Model} \label{subsec:kinematic_model}
We first develop a planar kinematic model for our electroadhesive auxetic layer stack to understand the relevant design variables. We chose to study square unit cells for their Poisson's ratio $\nu = -1$, which allows the sheet to retain geometric similarity to its original shape even when stretched \cite{Tang_Yin_2017, Jang_Won_Kim_Kim_Oh_Lee_Kim_2022}. A dimensional diagram of our square AUC is shown in Fig. \ref{fig:kinematics_dimension_drawing}. Each square has side length $a$ and out-of-plane thickness $t$, and the hinge joint between squares is modeled as a beam with width $\delta$ and length $c$. When stretched by an applied linear force $f$, each node reaches an angle $\theta$ with respect to the force axis, and each hinge joint reaches a radius of curvature $\rho$ and bending angle $\phi = 2\theta$. 


Inspired by \cite{Sedal_Memar_Liu_Menguc_Corson_2020}, we can estimate the AUC's equilibrium position $\theta$ by balancing the torques  about each individual square's center  from the applied force $\tau_f$, the bending of the hinge joints $\tau_j$, and the frictional torque from electroadhesion $\tau_{ea}$, as shown in Fig. \ref{fig:kinematics_dimension_drawing}(b). This model relies on three key assumptions: no out-of-plane motion, no buckling or plastic deformation, and that the hinge joints act as Euler-Bernoulli beams with no length change along their neutral axes.

 The applied force $f$ induces a torque about the center:
\begin{equation} \label{eq:torque_applied_force}
    \tau_{f} = fa(\cos(\theta) - \sin(\theta))
\end{equation}

As derived by \cite{Tang_Yin_2017}, we can model the hinge joints as Euler-Bernoulli beams. The bending radius and resulting torque from all four joints are thus:
\begin{equation} \label{eq:radius_cut_junction}
    \rho = \frac{c}{\phi}=\frac{c}{2\theta}
\end{equation}
\begin{equation} \label{eq:torque_cut_junction}
    \tau_j = \frac{4EI}{\rho} = \frac{2E}{3} \left(\frac{\delta^3 t}{c}\right) \theta
\end{equation}
where $E$ is the substrate's Young's modulus and $I = \frac{\delta^3 t}{12}$ is the hinge joint's area moment of inertia.

Finally, the normal force created from electroadhesion is:
\begin{equation} \label{eq:normalforce_electroadhesion}
    f_{ea} = \frac{\epsilon_0 \epsilon_r}{2} \left(\frac{V^2}{d^2}\right) A =: \sigma A
\end{equation}
where $\epsilon_0 = 8.854\cdot 10^{-12} N/V^2$ is the permittivity of free space, $\epsilon_r$ is the relative permittivity of the dielectric material, $V$ is the potential difference, $d$ is the dielectric thickness, $A$ is the overlap area of the charged electrodes, and $\sigma$ serves as the electrostatic pressure.

Assuming a frictional coefficient $\mu$ and, for simplicity, that the overlap area is a square of side length $a_{ov}$, we can then compute the frictional torque as:
\begin{align} \label{eq:torque_electroadhesion}
    \tau_{ea} &= \int r \mu \sigma dA \\
    &= \mu \sigma \int_{-a_{ov}/2}^{a_{ov}/2}\int_{-a_{ov}/2}^{a_{ov}/2} \sqrt{x^2 + y^2} \textrm{ } dx \textrm{ } dy \nonumber \\
    &= \frac{\mu \sigma a_{ov}^3}{6} \left(\textrm{arsinh}(1) + \sqrt{2}\right)\nonumber 
\end{align}

By balancing the torque equation
\begin{equation} \label{eq:torque_balance}
    \tau_f - \tau_j - \tau_{ea} = 0
\end{equation}
we can solve for the equilibrium angle $\theta$ via a root-finding algorithm such as Newton's method.

\subsection{Testing Setup for an Individual Auxetic Unit Cell} \label{subsec:sts2x2_testing_setup}
In order to test the performance of electroadhesion as a rotary friction mechanism for auxetic sheets, we patterned an auxetic unit cell into a flexible printed circuit board (PCB) (PCBWay, Shenzhen, China). The polyimide substrate is $t = 50$ \textmu m thick, and the AUC has dimensions $a = 17$ mm, $c = 3$ mm, and $\delta = 0.71$ mm. Because of the space taken up by wiring electrical traces, the total overlap area between copper electrodes on adjacent auxetic layers is 180.34 mm$^2$, for an average $a_{ov} = 13.43$ mm when unstretched.

We consider three dielectric films, which provide different trade-offs in price, relative permittivity, and thickness:
\begin{itemize}
    \item Aluminium-sputtered mylar (CS Hyde Company, Illinois, USA), 25 \textmu m thick mylar with a 25 nm Al layer, $\epsilon_r = 3$, cost = 0.00065 USD/cm$^2$ in bulk (4.6 m$^2$ rolls)
    \item Silver-sputtered PVDF-HFP (PolyK Technologies, LLC, Philadelphia, USA), 10 \textmu m thick biaxially oriented PVDF-HFP (90/10 wt) with 100 nm Ag layer, $\epsilon_r = 13$, cost = 0.0059 USD/cm$^2$ in bulk (68 m$^2$ rolls)
    \item Non-metallized PVDF-TrFE-CFE (PolyK Technologies, LLC), 22 \textmu m thick PVDF-TrFE-CFE (7 mol\% CFE), $\epsilon_r = 50$, cost = 0.19 USD/cm$^2$ in bulk (1.5 m$^2$ rolls)
\end{itemize}

Dielectrics and conductive adhesives were cut using a quasi-CW DPSS UV laser cutter (DPSS Lasers Inc., Series 3500). Dielectric films were cut into 17mm squares to match the auxetic pattern. For mylar and PVDF-HFP, we etched away their metallized sputter's 0.7mm outer ring to prevent arcing and adhered them to a 14mm square of 3M 1182 double-sided copper tape, before assembling them onto the flexible PCB as shown in Fig. \ref{fig:conceptual_diagram}. The non-metallized PVDF-TrFE-CFE was backed with 3M 9703 z-axis anisotropic conductive transfer tape before being cut and assembled.

\begin{figure}
    \centering
    \includegraphics[width=\columnwidth,clip]{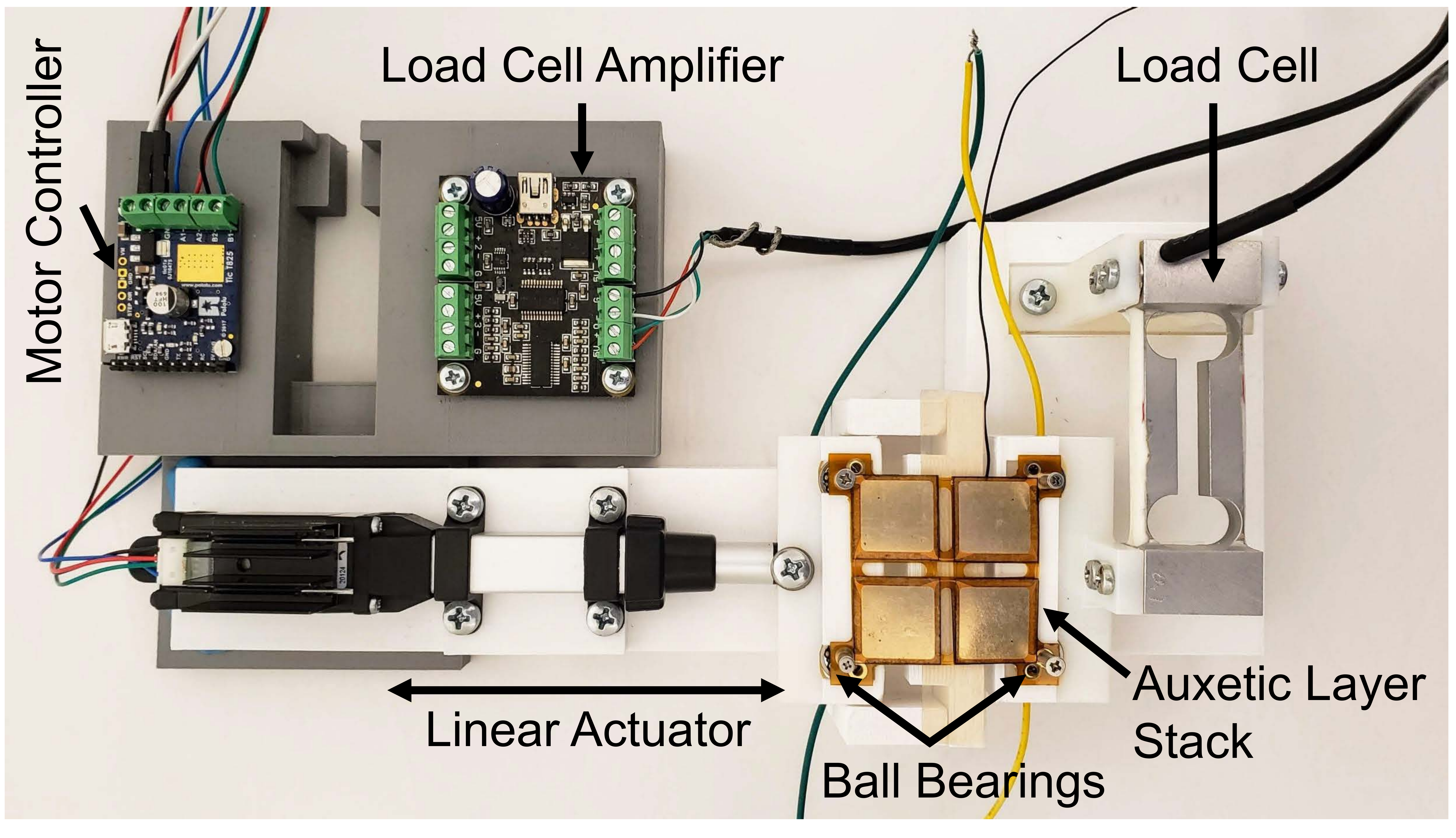}
    \caption{Testing setup for measuring the linear force vs. strain of an individual auxetic unit cell layer stack.}
    \label{fig:sts2x2_testing_setup}
\end{figure}

The testing setup, shown in Fig. \ref{fig:sts2x2_testing_setup}, measures hysteresis curves for the linear force required to stretch a multi-layer AUC stack under both locked and unlocked states. The flexible PCBs have four metallized mounting holes at the midpoint of their hinge joints. Each mounting hole is clamped between two threaded standoffs, which are connected to electrical ground and friction fit into ball bearings for free rotation as the AUC expands. The ball bearings are held in place vertically by a cover slip and given horizontal travel to expand when stretched, as expected from the auxetic structure's negative Poisson's ratio. All mounting components for the ball bearings are 3D printed with a glossy finish from VeroWhitePlus photopolymer using a Strasys Objet24 material jetting printer, ensuring a low coefficient of friction ($\mu = 0.34$). The hysteresis tests were conducted using an Actuonix S20 linear actuator, and linear force readings are measured at  4.9  Hz by a Loadstar Sensors RAPG-100G-A load cell. Because the tight space constraints make it difficult to actively actuate both the top and bottom AUCs at 90$^\circ$ angles relative to each other, we only actively stretch the bottom layer and let the other lie passively on top.

\subsection{Characterization Results for a Single Auxetic Unit Cell} \label{subsec:characterization_results_ets2x2}
\begin{figure}[!t]
    \centering
    \includegraphics[width=\columnwidth,clip]{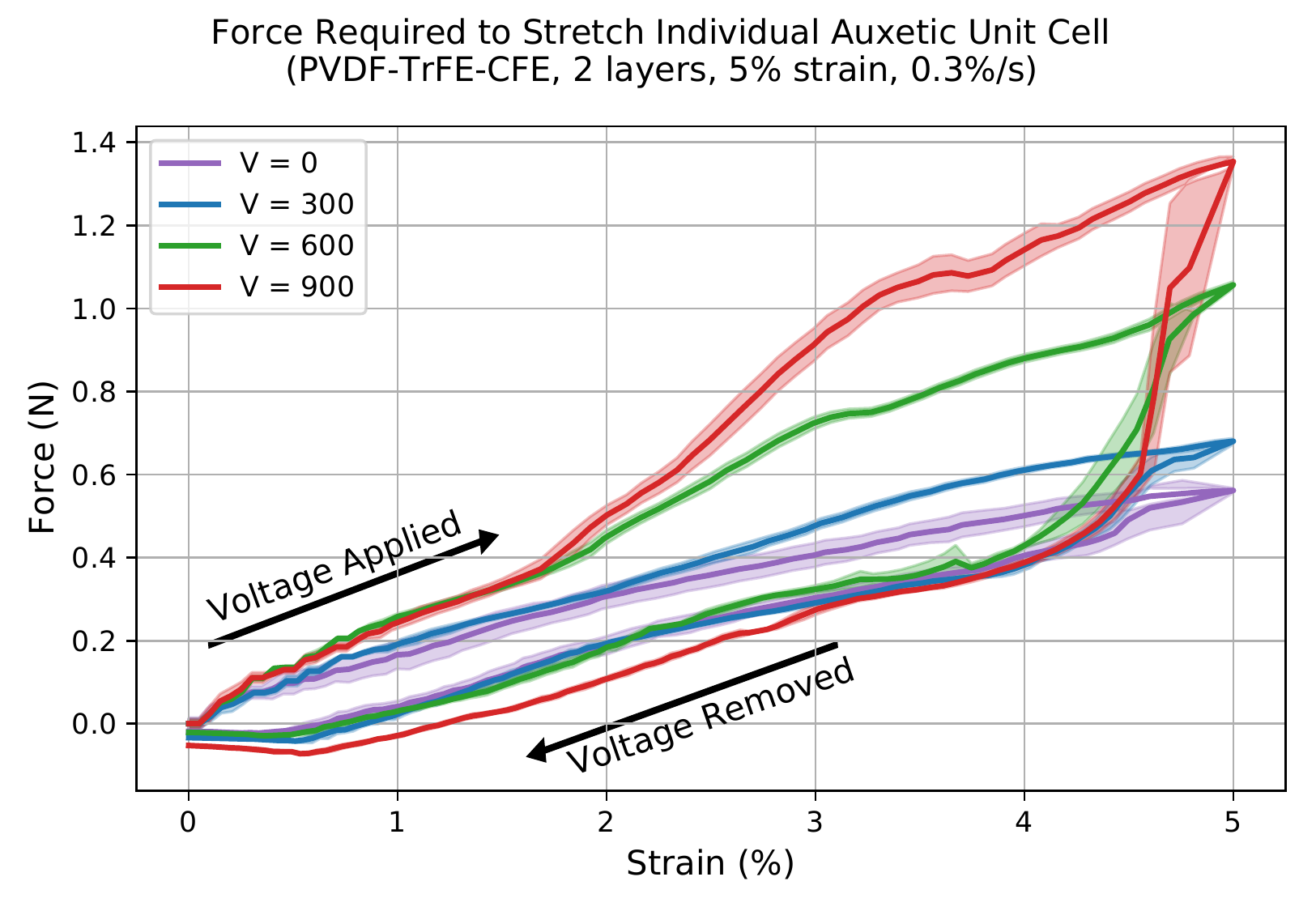}
    \caption{Hysteresis curve for the linear force required to stretch a 2-layer AUC stack with PVDF-TrFE-CFE dielectric. Voltage was applied only while stretching. The 95\% confidence interval is shown over $N = 4$ trials at each voltage.}
    \label{fig:ets2x2_force_vs_strain}
\end{figure}

Fig. \ref{fig:ets2x2_force_vs_strain} shows a sample hysteresis curve from the setup in Sec. \ref{subsec:sts2x2_testing_setup}. Testing was restricted to within 5\% strain to more accurately assess the effect of rotational friction without linear sliding, although higher voltages clearly show the slippage point around 3.5\% strain where the AUCs' locking ability gets outweighed by the load cell's spring constant of 1850 N/m.  The 2-layer PVDF-HFP dielectric film achieved the largest locked-over-unlocked stiffness variation of 7.6x with a locked linear stiffness of 1500 N/m, calculated as the average ratio of slopes across a 1\% strain range. It achieves this with a maximum power draw of 50 \textmu W/AUC, highlighting the low power consumption of electroadhesion relative to its locking force. Mylar and PVDF-TrFE-CFE achieved maximum stiffness variations of 4.6x and 7.0x, respectively. We proceed with PVDF-HFP for future experiments.

\begin{figure}[!th]
    \centering
    \includegraphics[width=\columnwidth,clip]{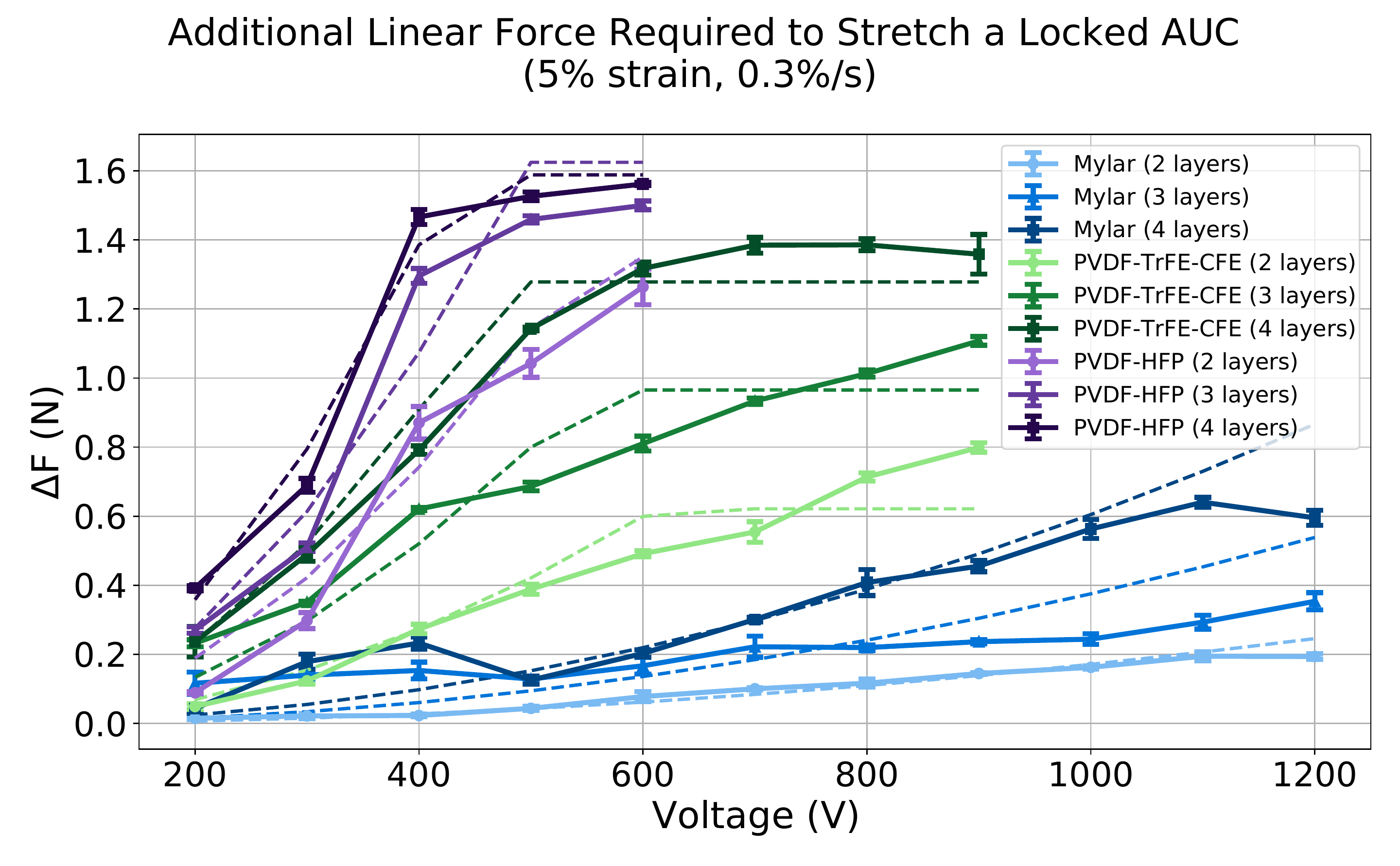}
    \caption{Additional linear force required to stretch a 2-, 3-, and 4-AUC stack when locked, as compared to when unlocked. Results are averaged over $N = 3$ trials, and the dotted lines show the fitted solid mechanics models.}
    \label{fig:ets2x2_force_vs_voltage_vs_numlayers}
\end{figure}

Fig. \ref{fig:ets2x2_force_vs_voltage_vs_numlayers} shows the additional linear force required to stretch a locked auxetic unit cell layer stack at 5\% strain, as compared to the unlocked layer stack. The solid mechanics model in Sec. \ref{subsec:kinematic_model} is plotted for comparison and generally matches the experimental trends. The solid mechanics model involves two fitted parameters. The Young's modulus $E$ is first fit using Brent's method to the unlocked AUCs' experimental data with an average fit of $E$ = 2.1 GPa ($\sigma$ = 0.4 GPa). Second, the effective frictional constant is fit to an average $\mu_{eff}$ = 0.13 ($\sigma$ = 0.14) to account for non-idealities in assembly that might introduce air gaps in between the dielectric and metallized electrodes. This Young's modulus and fitted efficiency match closely with values seen in the literature \cite{MCKEEN2017147, Zhang_Gonzalez_Guo_Follmer_2019, Hinchet_Vechev_Shea_Hilliges_2018}. Across all tests, the fitted solid mechanics model achieves a Root Mean Squared Error (RMSE) of 0.094 N. As shown by the horizontal predicted and experimental curves at high voltage, both PVDF-HFP and PVDF-TrFE-CFE achieve near-maximal locking in this testing apparatus.
\section{Characterization of 5x5 Auxetic Unit Cell Layer Stack} \label{sec:characterization_5x5}

\subsection{Assembly of a 5x5 AUC Layer Stack} \label{assembly_5x5}
\begin{figure}
    \centering
    \includegraphics[width=0.7\columnwidth,clip]{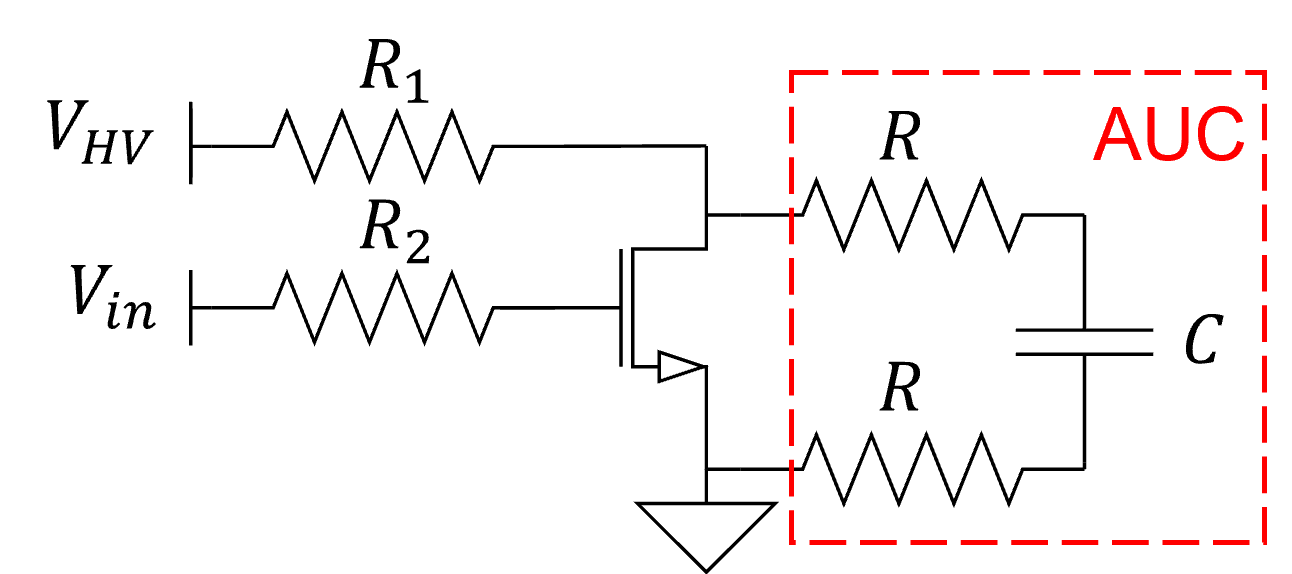}
    \caption{Logic circuit for driving an individual AUC. High voltage is supplied from a DC-DC voltage converter, and $V_{in}$ is obtained from a Teensy 4.1. AUCs can be modeled as capacitors $C \approx 8.3$ nF and two resistors $R \approx 1.2 \Omega$ in series.}
    \label{fig:nmos_source_follower}
\end{figure}

In order to explore how the results for an individual AUC extend to an auxetic array, we patterned a flexible PCB with a 5x5 array of square unit cells with the same geometry as in Sec. \ref{subsec:sts2x2_testing_setup}. Towards its functionality as a variable-stiffness surface for a formable crust shape display, we tested whether locking local subsets of AUCs can affect the array's global axial and transverse stiffness. We assembled the layer stack with PVDF-HFP as the dielectric, based on results in Sec. \ref{subsec:characterization_results_ets2x2}. Each flexible PCB includes through holes around the edges for easy mounting. Pin headers were soldered to the flexible PCBs and connected to ribbon cables, which were then routed to a logic PCB using the NMOS source follower circuit shown in Fig. \ref{fig:nmos_source_follower} for programmatic control of each auxetic unit cell via row-column addressing. A high voltage transistor (STMicroelectronics STB12NM60N) switches a row on one auxetic layer to either high voltage or ground, while another does the same to columns of the second layer. A high voltage DC-DC converter (EMCO F50) shifts applied voltages from 0.51 - 1.29 V to 200 - 600 V. A resistor ($R_1 = 10$ M$\Omega$) is used to limit current draw, and another resistor ($R_2 = 1$ k$\Omega$) limits burden on the Teensy 4.1 microcontroller supplying $V_{in}$. This high-voltage logic circuitry consumes a maximum of 72.2 \textmu A/AUC at 600 V. For portability, however, the below results were collected without this logic PCB by manually powering different electrode pairs.

\subsection{Axial Stiffness Variation of a 5x5 AUC Layer Stack} \label{subsec:characterization_5x5_linear}
In order to measure the ability of an auxetic layer stack to modulate its axial stiffness, we axially offset the dielectric-assembled auxetic sheet by half of an AUC relative to the second layer, and clamped the ends of both sheets at this offset using acrylic sheets and M2 screws through each flexible PCB's through holes. This causes the unit cells on subsequent layers to rotate in opposite direction just like when testing individual AUCs in Sec. \ref{subsec:sts2x2_testing_setup}, with the exception of only being able to lock 90\% of the electroadhesive pads at a time.

\begin{figure}
    \centering
    \includegraphics[width=\columnwidth,trim={10pt 0 0 0},clip]{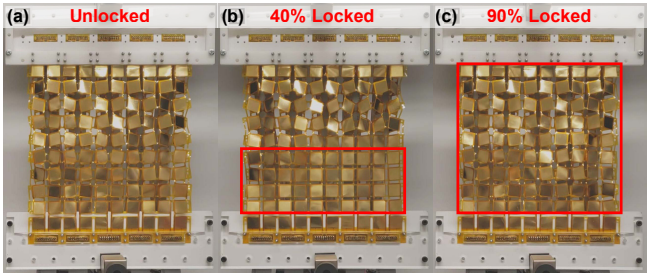}
    \caption{Example photos of the 5x5 AUC layer stack undergoing linear tests on an Instron at different locking conditions  at 5\% strain. Red boxes outline the locked cells.}
    \label{fig:instron_linear_pictures}
\end{figure}

\begin{figure}
    \centering
    \includegraphics[width=\columnwidth,trim={10pt 10pt 0 10pt}, clip]{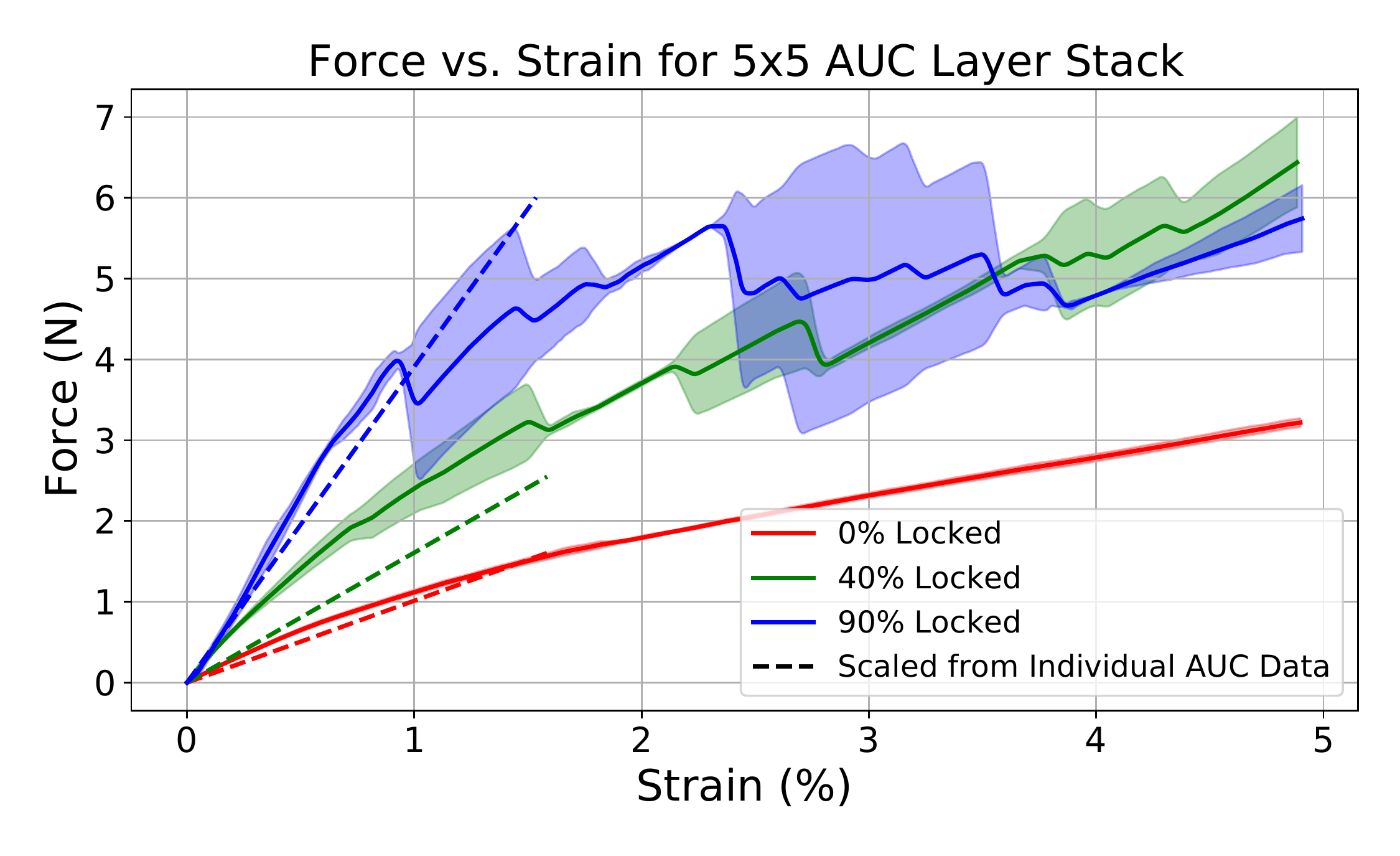}
    \caption{Linear force vs. strain for a 5x5 AUC layer stack under different locking conditions. Shaded regions show the 95\% confidence interval, averaged over $N = 3$ trials. High-uncertainty regions correspond to AUC slippage during individual tests. Dotted lines show the expected initial slope based on individual AUC data in Sec. \ref{subsec:characterization_results_ets2x2}.}
    \label{fig:instron_linear_force_vs_strain}
\end{figure}

We tested this multi-layer stack on an Instron 5560 with a 100 N load cell, with pictorial results shown in Fig. \ref{fig:instron_linear_pictures} and force vs. strain results shown in Fig. \ref{fig:instron_linear_force_vs_strain} for the cases when 0\%, the bottom 40\%, and all of the available 90\% of the rows were locked at 500 V. Tests were conducted up to 5\% strain with a 0.1\%/s strain rate. Scaled predictions in Fig. \ref{fig:instron_linear_force_vs_strain} were computed by treating the 5x5 cell array as a spring array with five columns in parallel and four springs in serial in each column (ignoring one AUC/column because of the attachment offset), and computing the entire array's axial stiffness based on the maximum 7.6x locked-over-unlocked stiffness ratio measured in Sec. \ref{subsec:characterization_results_ets2x2} for an individual AUC.

As shown in Fig. \ref{fig:instron_linear_force_vs_strain}, the unlocked case has the lowest stiffness and matches well with the predicted global axial stiffness. Unlike with individual AUCs, which had fixed boundary conditions at each corner, the full sheet is free to deform out-of-plane. The 90\% locked case's out-of-plane deformation in Fig. \ref{fig:instron_linear_pictures} highlights the necessity for some compliance when reaching the 5\% strain target, as shown by how its slope gradually decreases in Fig. \ref{fig:instron_linear_force_vs_strain} as AUCs slip and detach. Notably, both locked cases are much stiffer than predicted, which we hypothesize is because differences in how the two auxetic sheets deform out-of-plane actually press individual AUCs together, creating greater locking forces than the electroadhesion alone could produce. 

\subsection{Transverse Stiffness Variation of a 5x5 AUC Layer Stack} \label{subsec:characterization_5x5_rotary}
\begin{figure}
    \centering
    \includegraphics[width=\columnwidth,trim={10pt 0 0 0},clip]{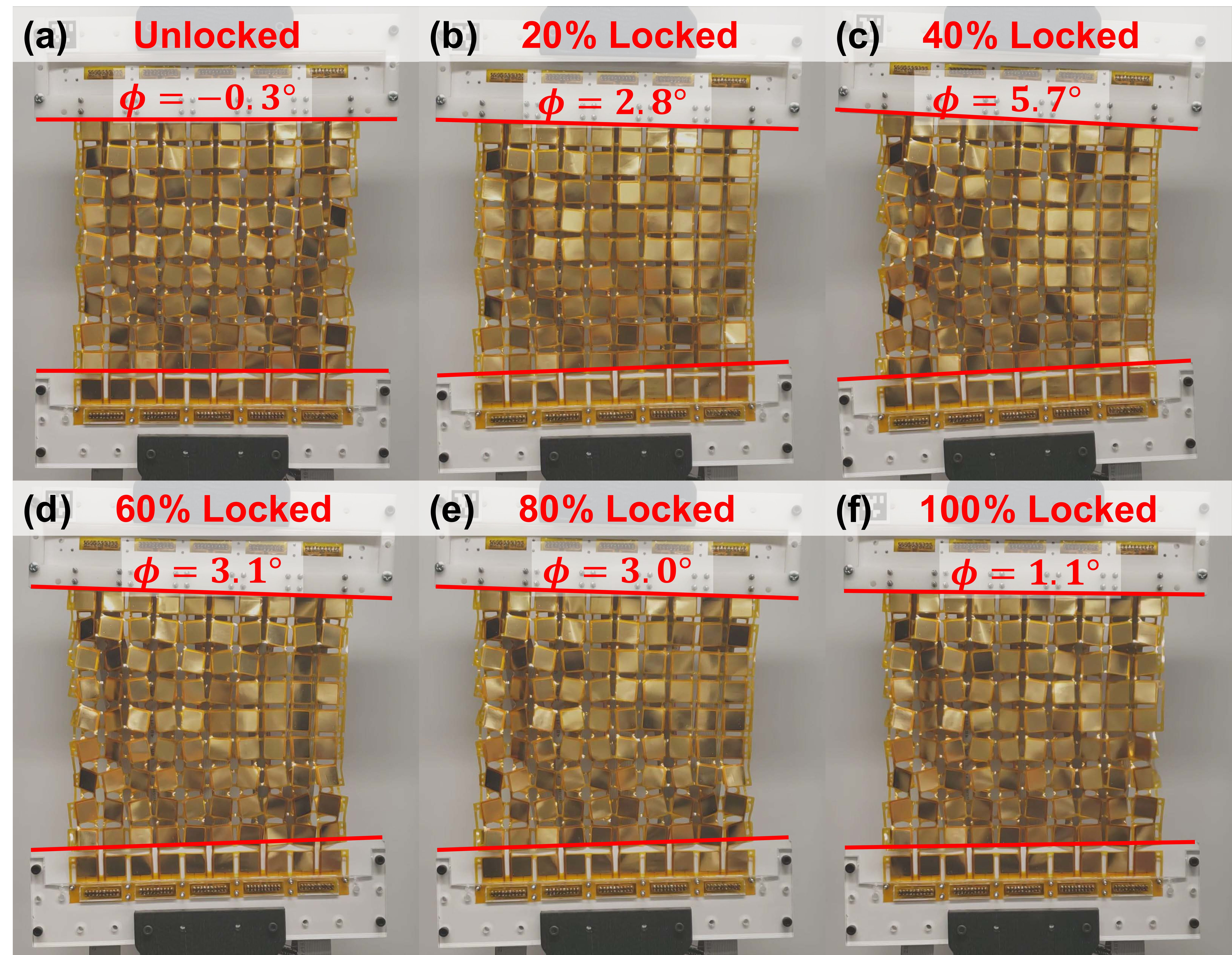}
    \caption{Example photos of the 5x5 AUC layer stack undergoing bending angle tests on an Instron at different locking conditions at 5\% strain. The total angle is computed by adding the rotations of the top and bottom mounts.}
    \label{fig:instron_rotary_pictures}
\end{figure}

\begin{figure}
    \centering
    \includegraphics[width=\columnwidth,trim={5pt 10pt 0 10pt},clip]{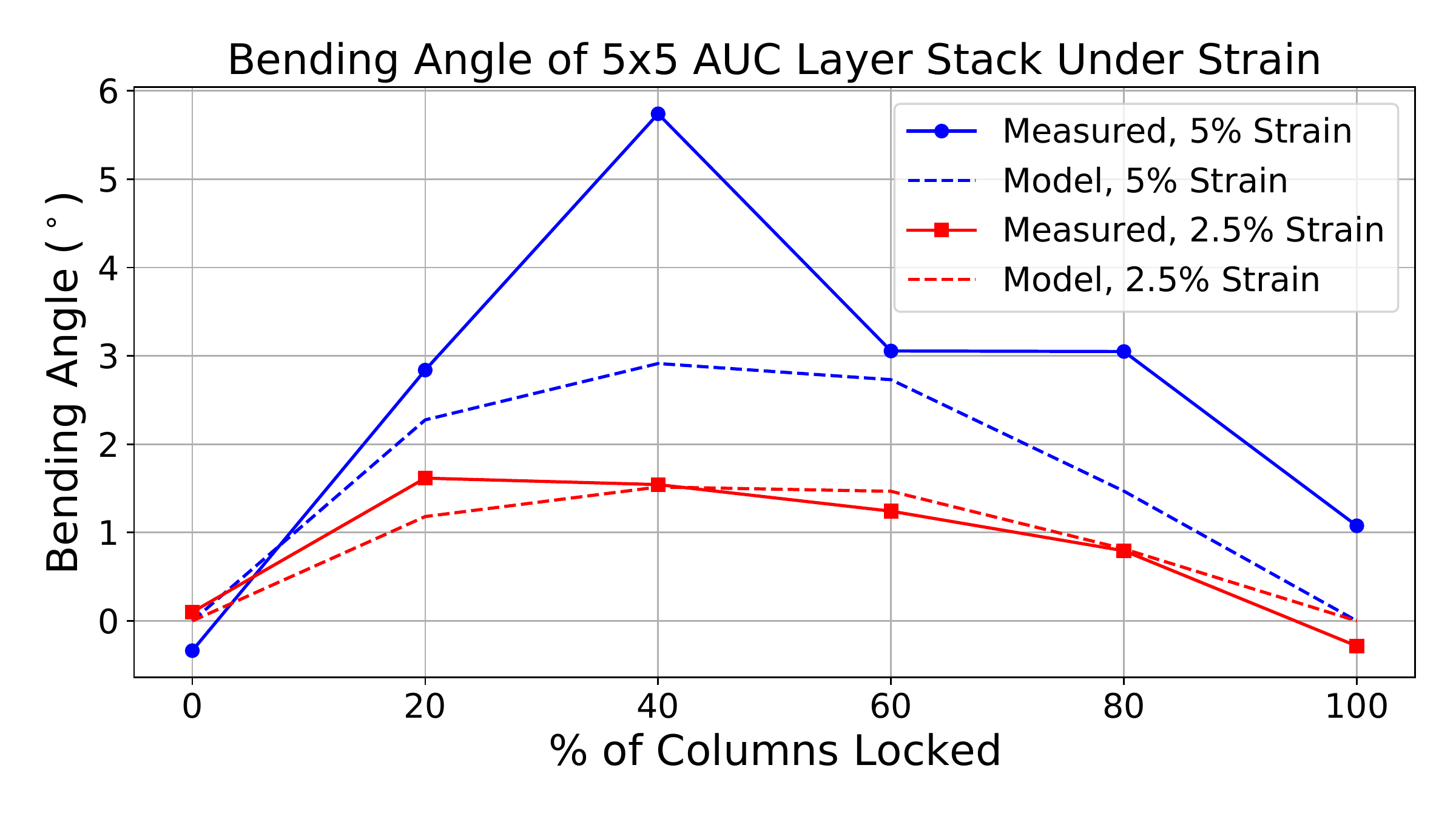}
    \caption{Bending angle vs. percentage of columns locked in a 5x5 AUC layer stack. Dotted lines show the expected bending angle based on individual AUC data in Sec. \ref{subsec:characterization_results_ets2x2}.}
    \label{fig:instron_rotary_bendingangle_vs_numcolumns}
\end{figure}

Unlike traditional uniaxial soft robotic applications, shape displays can make use of biaxial stiffness control to generate a more diverse range of shapes \cite{Shah_Yang_Yuen_Huang_Kramer_Bottiglio_2021}. Therefore,  we also tested the ability of this multi-layer stack to modulate its transverse stiffness and bend when given rotational freedom. The testing method was identical to Sec. \ref{subsec:characterization_5x5_linear}'s, except the mounting brackets were attached to ball bearings for free rotation and we varied the number of columns locked instead of rows. Results are shown in Fig. \ref{fig:instron_rotary_pictures} and Fig. \ref{fig:instron_rotary_bendingangle_vs_numcolumns}.

Fig. \ref{fig:instron_rotary_bendingangle_vs_numcolumns} highlights the influence of higher strain, and thus larger out-of-plane deformation, on the experimental data. Predicted results were obtained by solving an optimization problem for the minimum spring energy needed to position the 5x5 AUC spring array at the desired strain, based on the maximum 7.6x locked-over-unlocked stiffness ratio demonstrated in Sec. \ref{subsec:characterization_results_ets2x2} for an individual AUC. The experimental data matches the theory well for small strain, but similar to Sec. \ref{subsec:characterization_5x5_linear} the locking forces exceed predictions for higher strain. Future work should investigate these discrepancies further, but in Sec. \ref{sec:sdts_integration} we see that in practice mounting auxetic skins onto a soft robotic body creates boundary constraints that naturally limit out-of-plane deformation.

\section{Integration of Electroadhesive Auxetic Skins into a Formable Crust Shape Display} \label{sec:sdts_integration}

\begin{figure}
    \centering
    \includegraphics[width=\columnwidth,trim={2pt 0 0 0}, clip]{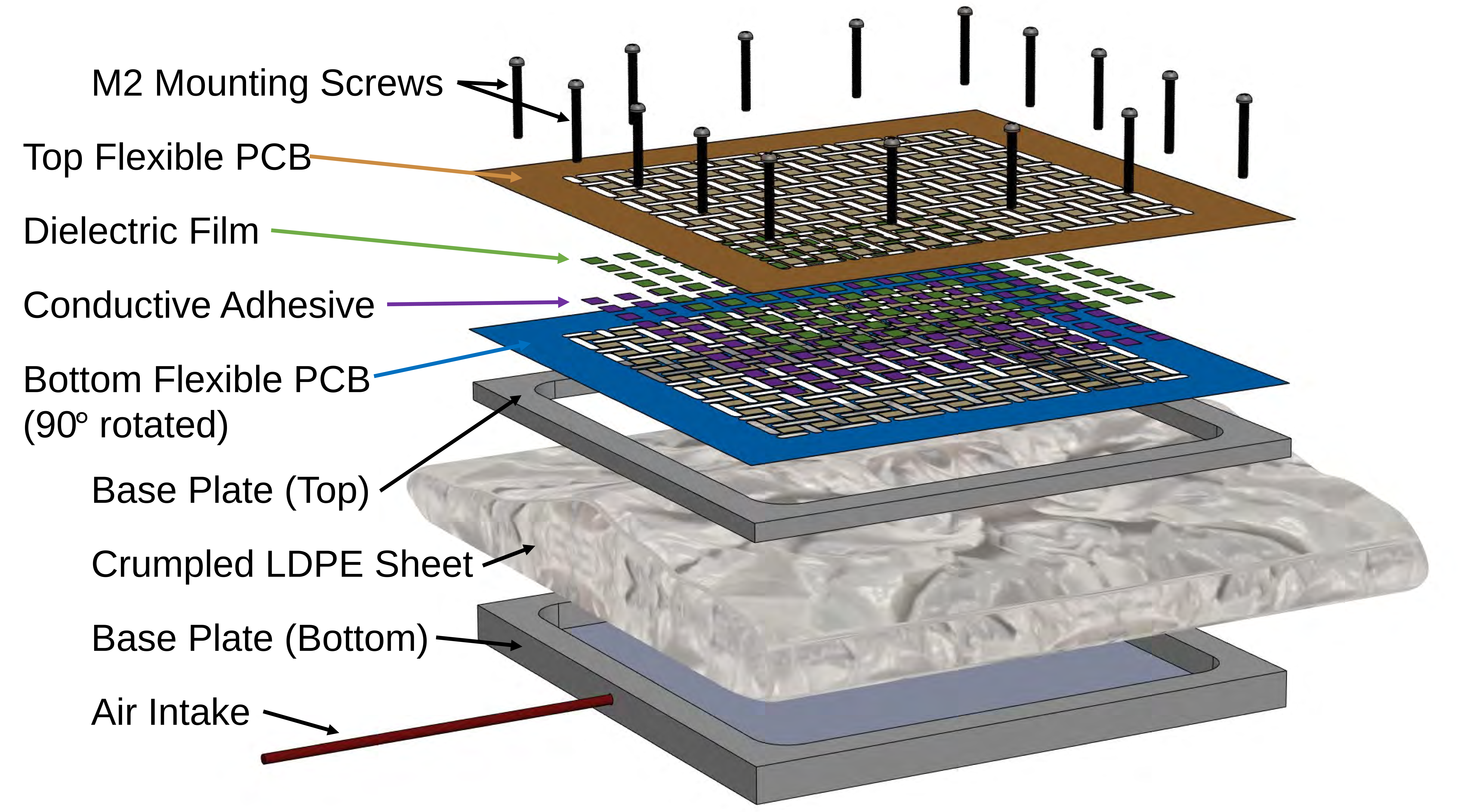}
    \caption{Full assembly stack of electroadhesive auxetic sheets onto an inflatable LDPE pouch to form a shape display.}
    \label{fig:sdts_layerstack}
\end{figure}

In order to test the performance of our electroadhesive auxetic skin for on-demand stiffness variation in a continuous shape display, we mounted it onto an inflatable LDPE pouch and measured its ability to modulate the shape output. Shown in Fig. \ref{fig:sdts_layerstack}, the inflatable pouch comprises a 50x50 cm$^2$ sheet of 30 \textmu m thick LDPE film crumpled to fit under the 20x20 cm$^2$ auxetic skin. The LDPE is locked at the edges between two laser cut acrylic sheets, and when pressurized air is pumped into the interior cavity the pouch balloons outwards into a uniform dome. The same 5x5 auxetic arrays used in Sec. \ref{sec:characterization_5x5} are mounted on top of the LDPE pouch, with the dielectric-assembled bottom sheet rotated by 90$^\circ$ relative to the top sheet. M2 mounting screws are threaded through the entire height of this assembly stack to fix the edges together.

\begin{figure}
    \centering
    \includegraphics[width=\columnwidth,clip]{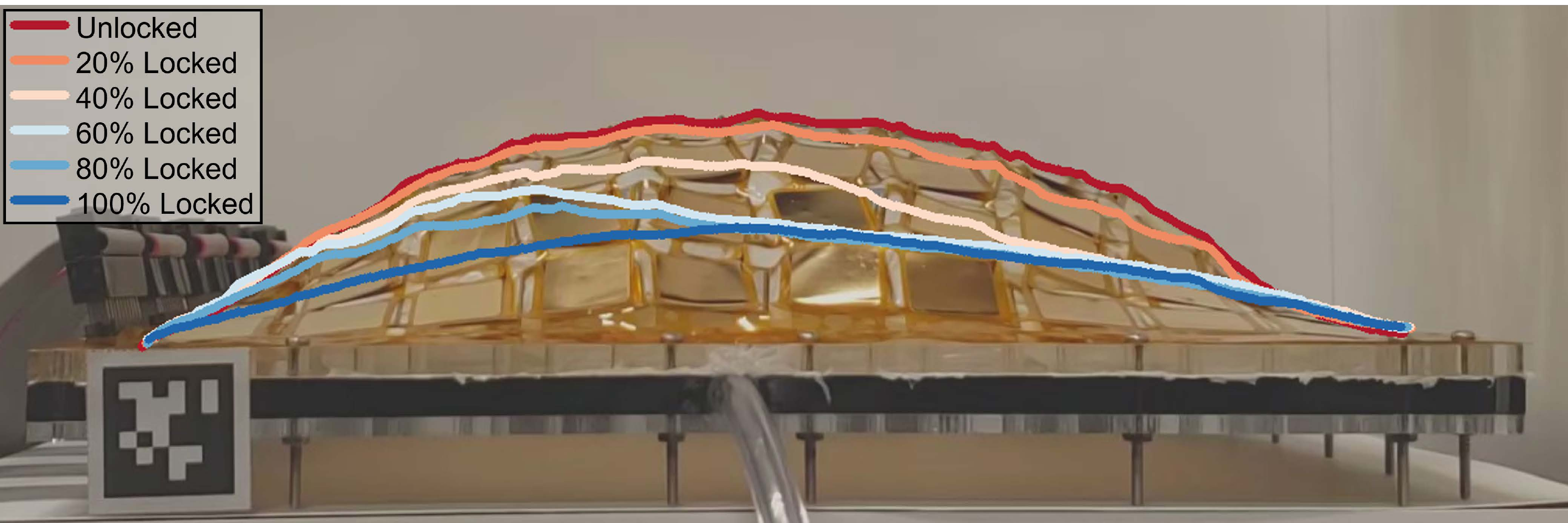}
    \caption{Side view of the shape display modulating its output shape based on the percentage of columns locked. Pressure within the LDPE pouch is held at 1 kPa.}
    \label{fig:sdts_sideview}
\end{figure}

\begin{figure}[!ht]
    \centering
    \includegraphics[width=\columnwidth,trim={20pt 3pt 1pt 22pt},clip]{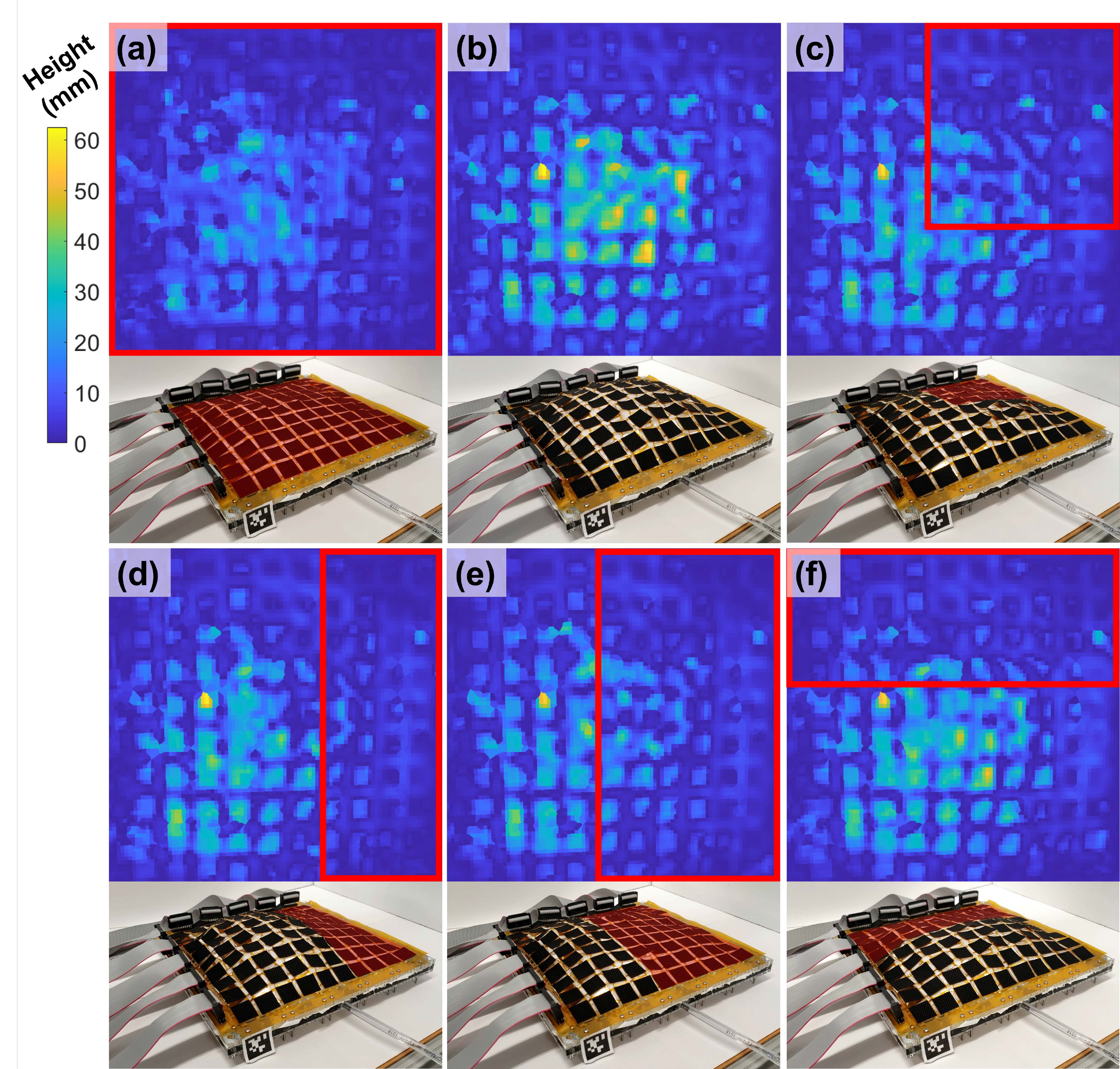}
    \caption{Elevation maps and perspective views of the shape display's output geometry under different locking conditions: (a) locking every cell, (b) locking no cells, (c) locking the top right 36\%, (d) locking the right 40\%, (e) locking the right 60\%, and (f) locking the top 40\%. Locked cells, outlined or shaded in red, expand noticeably less than unlocked cells. Heights are measured as the difference between a top-down Kinect's measured depth maps before and after inflation.}
    \label{fig:sdts_isometricview}
\end{figure}

Fig. \ref{fig:sdts_sideview} shows a side view of the electroadhesive auxetic skins modulating the output shape's curvature depending on the number of columns locked. Pressure is held at 1 kPa, which for the given assembly maximized inflation without electroadhesive slippage during any of the tested locking conditions. The auxetic skins enable a 2.3x decrease in global curvature between the unlocked and 100\% locked cases, and they achieve a 2.0x change in output slope between the display's left and right sides for the 80\% locked case. The LDPE pouch serves as a conformal substrate, limiting the out-of-plane deformation observed in Sec. \ref{sec:characterization_2x2}. Fig. \ref{fig:sdts_isometricview} further demonstrates how programmatically locking different sets of AUCs enables a single auxetic pattern to output a variety of global output shapes. Depth maps were recorded using a Microsoft Kinect for Xbox One, and black ripstop nylon squares were taped onto every AUC to limit reflectivity.

While data resolution still falls behind state-of-the-art \cite{Stanley_Okamura_2015, Zhang_Gonzalez_Guo_Follmer_2019} due to the inability to lock AUCs to the base plate, these results are promising for the application of electroadhesive auxetic skins as a programmable surface for formable crust shape displays, highlighting the ability of variable stiffness surfaces to actively control deformation and achieve shape change. The low cost (0.88 USD/AUC at quantities of 100 flexible PCBs, including high voltage logic circuitry), low power consumption (50 \textmu W/AUC), low profile ($<$300 \textmu m thickness), and monolithic manufacturing techniques for lowering assembly complexity decrease the barrier to entry traditionally required for distributed tactile displays, opening new options for fast prototyping and scalable deployment. 


\section{Conclusion and Future Work}

In this work, we motivate the application of electroadhesion as a layer jamming actuation mechanism for auxetic sheets and demonstrate its usage as a programmable strain limiting skin for a formable crust shape display. By using monolithic fabrication techniques such as flexible PCB manufacturing and laser cutting, these actuators present a scalable, low-cost, low-power, and easy to assemble approach to integrating variable stiffness capabilities into flexible robot systems. We demonstrated the ability of a multi-layer auxetic stack to programmatically modulate its axial and transverse stiffness, opening new possibilities in high degree-of-freedom control of soft robotic systems.

Our current design also has some drawbacks, notably in its simplified planar kinematic modeling and limited output shape change. Future work should investigate new analytical 3D models for electroadhesive auxetic skins, factoring in  plastic and out-of-plane deformation as well as fatigue. Additionally, further work is needed to optimize auxetic patterns supporting greater strain and thus increased global shape change. Future work should also consider locking shape display surfaces to the base plate for extra degrees of freedom \cite{Stanley_Okamura_2015}. Finally, other soft robotic applications such as vine robots \cite{Do_Banashek_Okamura_2020} and haptic clutches \cite{Hinchet_Vechev_Shea_Hilliges_2018} which also require low-profile, on-demand stiffness control should be explored. 

\addtolength{\textheight}{-2.7cm}   





\section*{Acknowledgments}

 This work is supported  by the National Science Foundation  (NSF)  Graduate Research Fellowship grant no. DGE-1656518, the  NSF  CAREER award grant no. 2142782, and the Alfred P. Sloan Research Fellowship grant no. FG-2021-15851. Part of this work was performed at the Stanford Nano Shared Facilities, supported by the   NSF  under award ECCS-2026822. We also acknowledge Kai Zhang, who ideated and prototyped early passive versions of this system.

\bibliographystyle{IEEEtran}

\end{document}